\documentclass[letterpaper]{article} 
\usepackage{aaai25}  
\usepackage{times}  
\usepackage{helvet}  
\usepackage{courier}  
\usepackage[hyphens]{url}  
\usepackage{graphicx} 
\usepackage{natbib}  
\usepackage{caption} 
\usepackage{algorithm}
\usepackage{algorithmic}

\usepackage{amsmath}
\usepackage{amssymb}

\usepackage[utf8]{inputenc}
\usepackage{enumitem} 
\usepackage{subcaption}

\usepackage{multicol}
\usepackage{multirow}
\usepackage{makecell}

\usepackage{booktabs}
\usepackage{tablefootnote}
\usepackage[table]{xcolor}
\usepackage{colortbl}      

\usepackage{newfloat}
\usepackage{listings}
\DeclareCaptionStyle{ruled}{labelfont=normalfont,labelsep=colon,strut=off} 
\floatstyle{ruled}
\newfloat{listing}{tb}{lst}{}
\floatname{listing}{Listing}
\title{DM-Adapter: Domain-Aware Mixture-of-Adapters \\ for Text-Based Person Retrieval}
\author{
    Yating Liu\textsuperscript{\rm 12}, Zimo Liu\textsuperscript{\rm 2}, Xiangyuan Lan \textsuperscript{\rm 24},
    Wenming Yang\textsuperscript{\rm 1},
    Yaowei Li\textsuperscript{\rm 23}\thanks{Corresponding author.},
    Qingmin Liao\textsuperscript{\rm 1*},
}
\affiliations{
    \textsuperscript{\rm 1}Shenzhen International Graduate School, Tsinghua University, China \\

    \textsuperscript{\rm 2} Pengcheng Laboratory,  China \\
    \textsuperscript{\rm 3} School of ECE, Peking University, China \\
    \textsuperscript{\rm 4} Pazhou Laboratory (Huangpu), China \\
    
   liuyatin21@mails.tsinghua.edu.cn, 
   \{liuzm, lanxy\}@pcl.ac.cn, \\
   yang.wenming@sz.tsinghua.edu.cn, ywl@stu.pku.edu.cn, liaoqm@tsinghua.edu.cn

}

\usepackage{bibentry}

\begin{document}

\maketitle

\begin{abstract}

Text-based person retrieval (TPR) has gained significant attention as a fine-grained and challenging task that closely aligns with practical applications. Tailoring CLIP to person domain is now a emerging research topic due to the abundant knowledge of vision-language pretraining, but challenges still remain during fine-tuning: (i) Previous full-model fine-tuning in TPR is computationally expensive and prone to overfitting.(ii) Existing parameter-efficient transfer learning (PETL) for TPR lacks of fine-grained feature extraction. To address these issues, we propose \textbf{D}omain-Aware \textbf{M}ixture-of-\textbf{Adapters} (\textbf{DM-Adapter}), which unifies Mixture-of-Experts (MOE) and PETL to enhance fine-grained feature representations while maintaining efficiency. Specifically, \textbf{Sparse Mixture-of-Adapters} is designed in parallel to MLP layers in both vision and language branches, where different experts specialize in distinct aspects of person knowledge to handle features more finely. To promote the router to exploit domain information effectively and alleviate the routing imbalance, \textbf{Domain-Aware Router} is then developed by building a novel gating function and injecting learnable domain-aware prompts. Extensive experiments show that our DM-Adapter achieves state-of-the-art performance, outperforming previous methods by a significant margin.

\end{abstract}



\section{Introduction}

Text-based Person Retrieval (TPR) \cite{Li_2017_CVPR} is a cross-modal task that aims to retrieve persons from a large-scale image pool based on textual descriptions instead of images, which is crucial in intelligent transportation and security scenarios, especially when witnesses can only provide textual descriptions without any target images.

Due to the remarkable generalization in  cross-modal understanding, Vision-Language Pre-training (VLP) models \cite{zhang2024vision} have garnered extensive interest from both academia and industry.
Among these, the most representative work is Contrastive Language-Image Pretraining (CLIP) \cite{clip}, which was pretrained on a large-scale dataset of 400 million image-text pairs.
FFT (Fully Fine-Tuning)-based methods, such as IRRA \cite{jiang2023cross} and CFine \cite{yan2023clip}, train the entire model and leverage CLIP's original knowledge only during initialization.
These methods typically employ complex fine-grained modules to enhance performance. 
In contrast, the PETL-based mechanism like CSKT \cite{liu2024clip} freezes the whole backbone of CLIP and introduces effective components such as bidirectional vision-language prompts to fine-tune a small number of parameters, which achieves comparable results while significantly reducing training and storage costs.
\begin{figure}[t]
    \centering
    \includegraphics[width=1.0\linewidth]{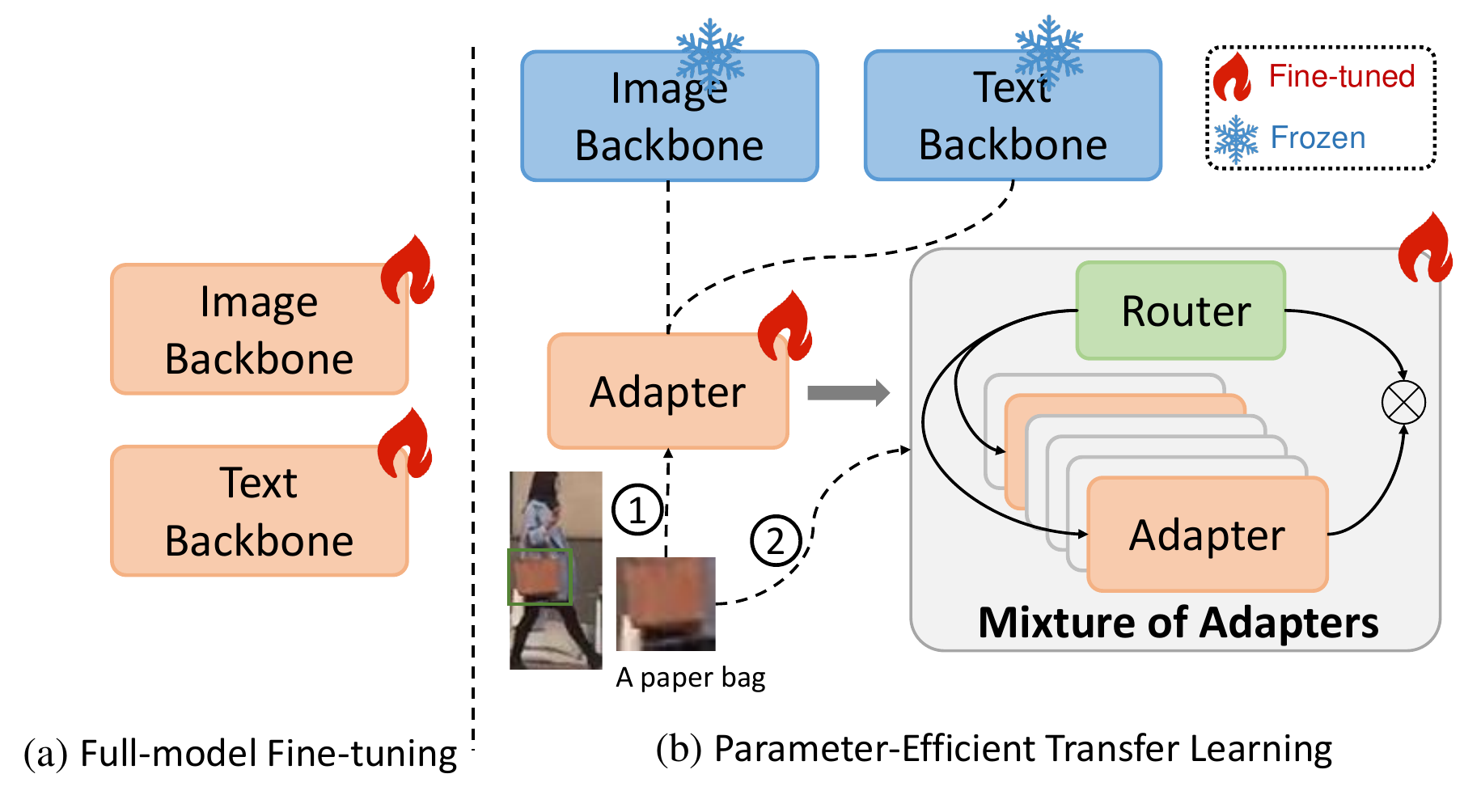}
    \caption{\normalsize{Evolution of CLIP-based paradigms for text-based person retrieval. (a) The FFT-based method unfreezes and trains the entire model. (b) The recent PETL-based method freezes CLIP and  uses a single adapter on the input token as shown in the \emph{left}. 
    Our mixture-of-adapters achieves the fine-grained knowledge transferring with MOE in the \emph{right}.} }
\label{fig:evo}
\end{figure}

\begin{figure}[t]
    \centering
    \includegraphics[width=0.92\linewidth]{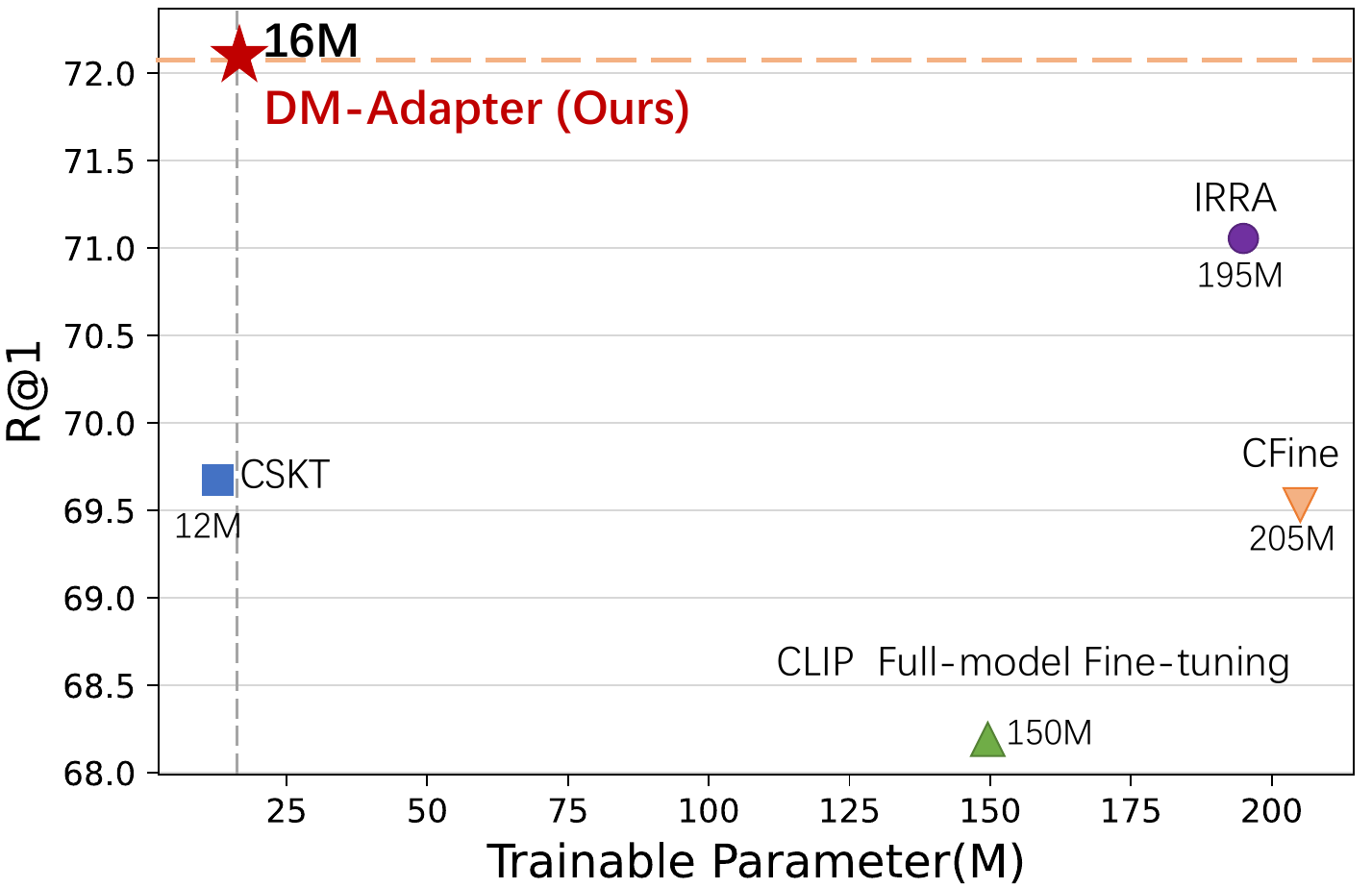}
    \caption{\normalsize{Comparison with CLIP-based methods. Our approach achieves the best trade-off between performance and parameter efficiency.} }
\label{fig:compare}
\end{figure}

However, existing approaches still face two primary challenges: (i) FFT-based methods, despite their fine-grained transfer capabilities, require enormous time and computational consumption, and have a risk of overfitting on relatively small-scale person datasets.
(ii) Recent PETL-based method CSKT, while efficient in trainable parameters, lacks of more fine-grained and specialized considerations for the specific domain of person retrieval, leading to a decline in overall performance. 
Meanwhile, the success of mixture-of-experts (MOE) models \cite{jiang2024mixtral} in foundation models that expands a feedforward block into multiple blocks, offers the potential of designing fine-grained PETL-related components.



In this paper, we propose \textbf{DM-Adapter}, \emph{i.e.}, \underline{\textbf{D}}omain-Aware \underline{\textbf{M}}ixture-of-\textbf{Adapters} for Text-Based Person Retrieval to achieve robust performance while keeping parameter efficiency. 
To make the vanilla adapter more fine-grained and specialized, we first design vision-language \textbf{Sparse Mixture-of-Adapters (SMA)}, which are composed of a Top-K router and multiple adapters spanning MLP layer.
SMA enables different adapter experts to specialize in distinctive aspects of person characteristics, thereby achieving fine-grained transfer learning.
Meanwhile, to alleviate the routing imbalance in MOE and incorporate domain information related to person retrieval, \textbf{Domain-Aware Router (DR)} is proposed to establish a coupling relationship between domain information and router by designing a domain-aware gating function and injecting learnable prompts to the gate, which helps our model to select expert adapters more effectively.
\renewcommand{\thefootnote}{}
\footnotetext{\raggedright Code: \url{https://github.com/Liu-Yating/DM-Adapter}}
\renewcommand{\thefootnote}{\arabic{footnote}}

In overall, DM-Adapter transfers the person-related knowledge of CLIP effectively based on mixture-of-experts and parameter-efficient transfer learning.
As illustrated in Figure \ref{fig:compare}, our approach surpasses previous state-of-the-art methods while training only 16M parameters.

Our contributions are summarized as follows:
\begin{itemize}
    \item To our knowledge, our research is the first to explore a MOE framework based on PETL for TPR, which implicitly mines fine-grained person knowledge without requiring any additional complex interaction modules.

    \item We design a novel domain-aware router by incorporating domain information by several learnable prompts to alleviate the imbalanced routing issue.

    \item We conduct comprehensive experiments on three public benchmarks \emph{i.e.}, CUHK-PEDES, ICFG-PEDES and RSTPReid, and the results demonstrate the superiority of the proposed DM-Adapter framework.

\end{itemize}

\section{Related Work}
\subsection{Text-based Person Retrieval}

Text-based Person Retrieval (TPR) along with the benchmark dataset CUHK-PEDES, was first proposed by Li \emph{et al.} \cite{Li_2017_CVPR} to solve the problem that the target query images are not always available.
Earlier research predominantly adopted separate uni-modal backbones \cite{shu2022see,farooq2022axm} including ResNet, ViT, LSTM or BERT to extract the vision and language features, and their representations are then aligned by global \cite{zhang2018deep,chen2022tipcb} or local \cite{ding2021semantically,gao2021contextual} matching methods.
Ding \emph{et al.} proposed a cross-modal implicit relation reasoning and aligning framework based on Contrastive Language–Image Pre-training (CLIP) \cite{clip}.
Yan \emph{et al.} \cite{yan2023clip} also designed a multi-grained matching method based on CLIP to mine cross-modal correspondences from coarse to fine.
CLIP containing abundant vision and language knowledge simultaneously has emerged as a key backbone in TPR \cite{song2024diverse,zhao2024unifying,liu2024causality,cao2024empirical}.
Liu \emph{et al.} \cite{liu2024clip} first developed a novel parameter-efficient transfer learning method CSKT based on CLIP for TPR, which outperforms the performance of full-tuning CLIP with only fine-tuning 12M parameters.
This motivates us to further explore how to adapt CLIP to text-based person retrieval with fewer parameters for more efficient and effective performance.

\subsection{Parameter-Efficient Transfer Learning}
Parameter-Efficient Transfer Learning (PETL) \cite{han2024parameter} provides a modular and efficient way to address the challenges of full-model fine-tuning (FFT) such as catastrophic forgetting and high computation costs. 
They work by keeping most of the pretrained model's weights fixed, only updating a small subset of parameters to efficiently adapt the model to a new task.
Adapter \cite{houlsby2019parameter} and Prompt \cite{liu2021p} were introduced to facilitate the transfer of large language models to specific downstream tasks by inserting \emph{additional} parameters to models.
LoRA  \cite{hu2021LoRA} as a \emph{reparameterized} fine-tuning methods, utilizes low-rank decomposition to reconstruct the weight matrices.
Subsequently, cross-modal prompt MaPLe \cite{khattak2023maple} were proposed in vision-language and further achieve cross-modal interactions. 
In CSKT \cite{liu2024clip}, PETL was first successfully incorporated in CLIP for text-based person retrieval, which designed bidirectional prompts and dual-branch adapters to achieve superior performance compared to FFT.
However, it still relies on fundamental PETL configurations, and CLIP-based PETL has not yet reached its limits in TPR.

\subsection{Mixture-of-Experts}
Mixture-of-Experts (MOE) has been extensively explored in computer vision \cite{riquelme2021scaling}, natural language processing \cite{shazeer2017outrageously} and vision-language pretraining \cite{chen2024eve}, which designs multiple separate experts to scale up models, and integrates a gate function to modulate the contributions of each expert.
Sparse Mixture-of-Experts (SMoE) \cite{jiang2024mixtral} has recently gained widespread attention in large language models, which strategically activates distinct experts for input via a router, thereby yielding noteworthy efficiency enhancements.
Despite the powerful capabilities of MoE, there is still rare exploration about MOE in transfer learning for cross-modal downstream tasks. 

\begin{figure*}[t]
	\centering
		\includegraphics[width=0.85\linewidth]{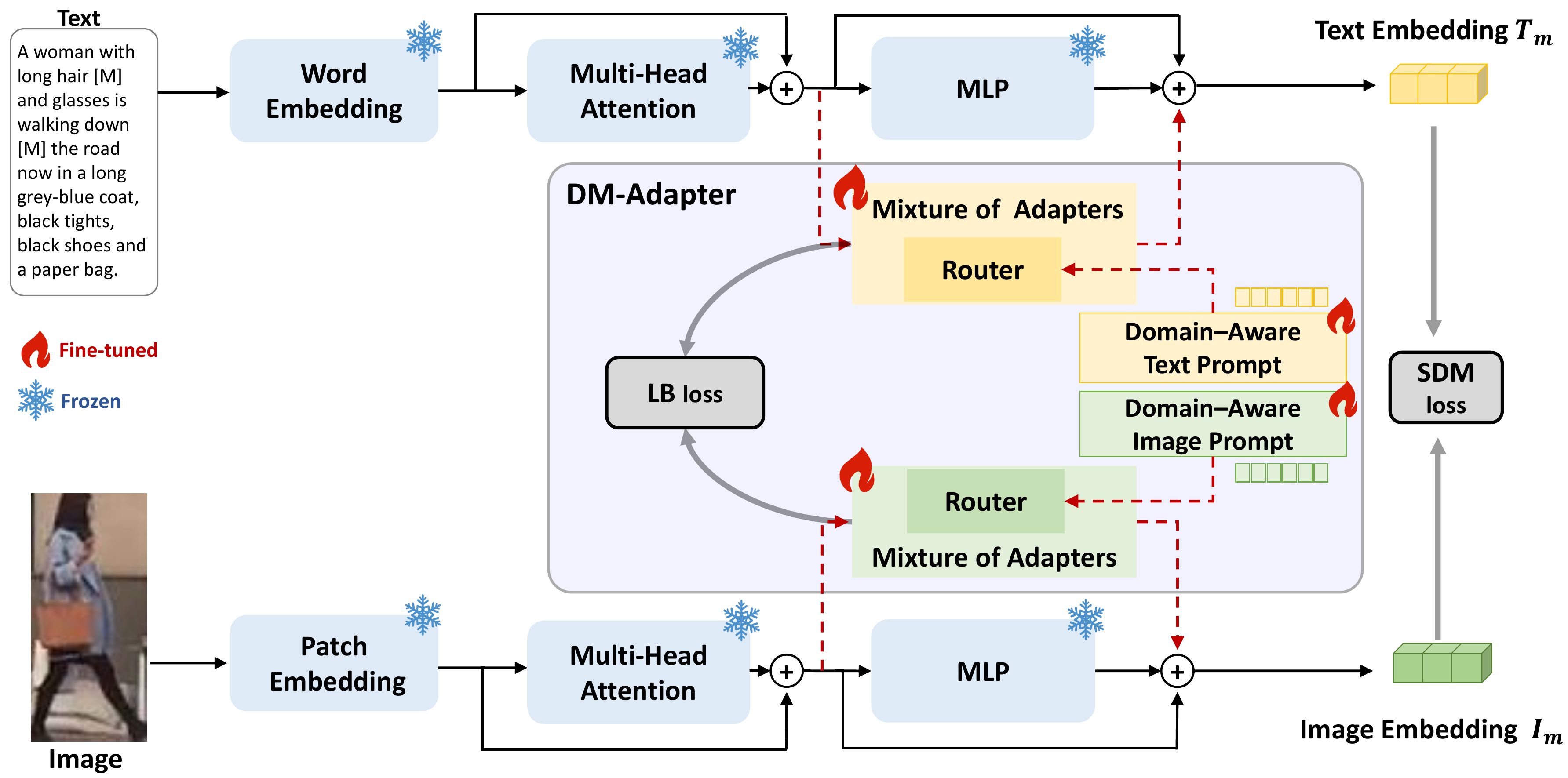}
	\caption{The overall framework of the proposed method. We adopt CLIP (ViT-B/16) as backbone, and design Domain-Aware Mixture-of-Adapters spanning MLP layer. The full parameters of vanilla CLIP are frozen during training phase. Only a fewer of parameters in DM-Adapter are trainable. The overall optimization objective incorporates SDM loss and LB auxiliary loss.}
	\label{fig: framework}
\end{figure*}

\section{Methodology}
\subsection{Framework}
As show in Figure \ref{fig: framework}, we adopt CLIP (ViT-B/16) as our backbone network and design Domain-Aware Mixture-of-Adapters (DM-Adapter) spanning MLP layers in both vision and language branches, which is capable of transferring knowledge within CLIP for TPR  with only fine-tuning a small amount of parameters. 
Each DM-Adapter consists of a mixture-of-adapters and a domain-aware router to generate more fine-grained and specialized representations, where an auxiliary loss is incorporated to balance the load of the router.

For image encoder, the input image $I$ is first partitioned to a sequence of $N$ non-overlapping patches.
The patches are then mapped to embeddings with a linear projection and added with positional embeddings to enhance spatial information. 
Subsequently, a [CLS] token is introduced at the beginning of the embeddings to denote the overall global representation of the image. 
The sequence of $N^2+1$ tokens is then fed into a series of transformer blocks, where a transformer block typically consists of a Multi-Head Attention (MHA) and a MLP. 
Layer normalization is omitted for simplicity in the framework.
We finally obtain visual representations $\{v_{cls}, v_1,\cdots,v_N\}$ with $v_{cls}$ being the global visual representation. 

For text encoder, the input description $T$ is tokenized to embeddings $f$ by a simple tokenizer with a 49152 vocab size.
$f$ then adds [BOS] as the start of the sequence and [EOS] as the end flag.
Thus, the overall sequence can be denoted as $\left\{ {{f_{bos}},{f_1}, \ldots ,{f_{eos}}} \right\}$ and then fed into the transformer as above image encoder,
where the output of $f_{eos}$ is the global representation in language branch.

The visual representation $v_{cls}$ and textual representation $f_{eos}$ are finally interacted and calculated by Similarity Distribution Matching (SDM) \cite{jiang2023cross}  which is an effective matching loss function across different modalities. 

\subsection{DM-Adapter}

\subsubsection{Motivation \& Intuition.}
Adapter \cite{chen2022adaptformer} as the most popular PETL approach, has demonstrated both its effectiveness and efficiency in fine-tuning various vision and language large models, which forms the foundation of our method.
It inserts small modules into transformer layers, which employs a down-projection ${\boldsymbol{W}_{{\rm{down }}}} \in {{\mathbf R}^{d \times m}}$ to map the input $x$ to a lower-dimensional space defined by the bottleneck dimension $m$, followed by a nonlinear activation function $f$ like ReLU and an up-projection with ${\boldsymbol{W}_{{\rm{up}}}} \in {{\mathbf R}^{m \times d}}$.
Adapter is then incorporated with a residual connection, formulated as:
\begin{equation}
h' \leftarrow h+f\left(x \boldsymbol{W}_{\text {down }}\right) \boldsymbol{W}_{\text {up }},
\label{eq:adapter}
\end{equation}
where $h$ is the output of the original $x$, and $h'$ represents the final output with a adapter.

For the fine-grained TPR task, merely relying on a single adapter to extract comprehensive features remains challenging. 
The recent works like sparsely-activated mixture-of-experts (MoE) models \cite{jiang2024mixtral} have been shown effective in LLMs by merging and activating only a subset of FFN layers for each input.
This approach is based on the assumption that a larger model capacity allows for the accommodation of more information.   
Thus, it offers the potential for fine-grained feature representation by leveraging MOE, where each expert can handle different aspects of the features, enabling more detailed representations.
\begin{figure}[!tb]
	\centering
		\includegraphics[width=0.99\linewidth]{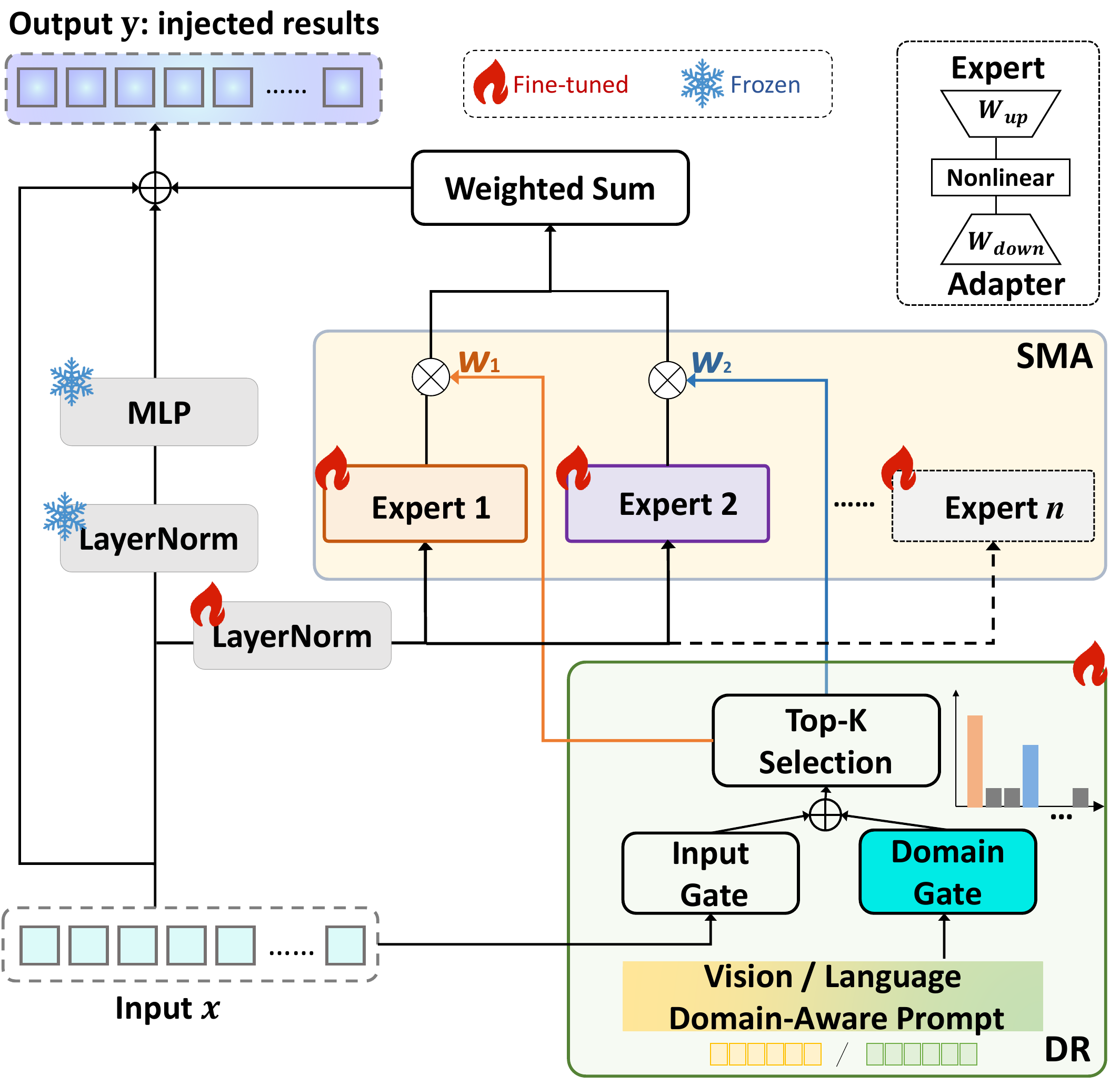}
	\caption{\normalsize{Architecture of DM-Adapter. DM-Adapter is mainly composed of SMA and DR. DR inserts novel domain-aware prompts on the input tokens, and designs an domain gating to capture these prompts.}}
	\label{fig: moe}
\end{figure} 

\subsubsection{Overview of DM-Adapter.} 
To this end, in Figure \ref{fig: moe}, we first design \textbf{S}parse \textbf{M}ixture of \textbf{A}dapters (\textbf{SMA}) in parallel with the MLP layers of each transformer. 
This architecture leverages the non-linear feature transformations of MLP layers to enhance the model's adaptability by integrating sparse mixture-of-adapters.
The output of SMA for a given input $x$ is determined by the weighted sum of the selected adapters, where the weights are given by the gating network $G$ of a router as shown in Equation (\ref{eq:ori_sma}).
Gating in Equation (\ref{eq:gate1}) enhances the information capacity of SMA while decreasing computation costs.
However, the general router process input information in a fully data-driven manner, lacking of any prior knowledge specific to person retrieval.
Therefore, we further design a novel domain-aware router that embeds specific domain-aware information into the original router by injecting learnable prompts, the final output is refined in Equation (\ref{eq:total}.)
In addition, to ensure the effectiveness of our designed domain-aware router, a load-balancing loss is further developed to balance load of experts.

\subsubsection{Sparse Mixture-of-Adapters.}
To enable fine-grained representations, given $n$ expert adapters, the forward process to MLP layer with SMA can be expressed as:
\begin{equation}
\sum\limits_{i = 0}^{n - 1} G {(x)_i} \cdot {Adapter_i}(x),
\label{eq:ori_sma}
\end{equation}
where $G {(x)_i}$ is $n$-dimensional output representing the gating weight for the $i$th adapter as equation (\ref{eq:adapter}).
We adopt a simple and effective gating mechanism \cite{jiang2024mixtral} by taking the softmax over the Top-K logits of a linear layer:
\begin{equation}
G(x)=\operatorname{Softmax}\left(\operatorname{TopK}\left(x \cdot W\right)\right),
\label{eq:gate1}
\end{equation}
where Top-K refers to selecting the highest $K$ weights, $W$ denotes the gating function weight of input tokens, and softmax performs normalization of the selected logits.
This sparse mechanism ensures that if the $K$ is fixed, the model's capacity is enhanced with the increase of $n$ while its computation costs remain consistently stable.
Considering the above, the output $y$ after a MLP layer and $n$ expert adapters for an input $x$ is formulated as:
\begin{equation}
\begin{array}{l}
\begin{array}{*{20}{c}}
{}&{}&{\begin{array}{*{20}{c}}
{}&{}
\end{array}}
\end{array}h_o = x + MLP(LN(x))\\
y = h_o + \sum\limits_{i = 0}^{n - 1} {{\mathop{\rm Softmax}\nolimits} } {\left( {{\mathop{\rm TopK}\nolimits} \left( {x \cdot{W}} \right)} \right)_i} \cdot { Adapter_i}(x),
\end{array}
\label{eq:total}
\end{equation}
where $h_o$ is the original output of MLP,  and $y$ is the final output with expert adapters.

\subsubsection{Domain-Aware Router.}
The general gating function is typically determined by the input tokens like Equation (\ref{eq:gate1}) and (\ref{eq:total}).
Here, MOE places no prior constraints on the router, which can easily result in imbalanced routing.
Meanwhile, existing routing ignores domain information when transferring foundation models to the specific TPR task.

To supplement domain-specific person knowledge, we propose a \textbf{D}omain-Aware \textbf{R}outer (DR), which incorporates domain information by several tunable prompts $p$ embedded in vision and language, and designs a domain-aware gating function $p\cdot W_d$ based on these prompts.  
Thus, the item $x\cdot W$ in Equation (\ref{eq:gate1}) and  (\ref{eq:total}) is then modified to $x\cdot W + p\cdot W_d$.
The output of DM-Adapter can be reformulated as:

\begin{equation}
\resizebox{0.47\textwidth}{!}{$
y = {h_o} + \sum\limits_{i = 0}^{n - 1} \text{Softmax} \left( \text{TopK}\left( x \cdot W + p \cdot {W_d} \right) \right)_i \cdot Adapter_i(x)
$}.
\end{equation}

\subsubsection{Load-Balancing loss.}
To ensure the effectiveness of our designed domain-aware router, an auxiliary loss is utilized to balance routing in mixture-of-adapters, which encourages experts to receive roughly equal numbers of training samples, and avoids concentrating the load on a single expert. 
Inspired by \cite{chen2024eve} and \cite{shazeer2017outrageously}, we design Top-K \textbf{L}oad-\textbf{B}alancing (\textbf{LB}) loss to balance the average weights of the selected K experts:
\begin{equation}
\mathcal{L}_{a u x}=\alpha \cdot \sum_i^n f_i \times p_i,
\label{eq:aux_loss}
\end{equation}
where $f_i$ is the fraction of tokens assigned to the $i$th expert under the Top-K mechanism, $p_i$ represents the average routing weight for the $i$th expert, and $\alpha$ is a hyperparameter.


\subsection{Optimization and Inference}
A parameter-free loss function is adopted in training phase termed as Similarity Distribution Matching (SDM) \cite{jiang2023cross}, which integrates cosine similarity distributions of the $N\times N$ embeddings for image-text pairs into the KL divergence to build up the connection of two modalities.

\begin{equation} 
\mathcal{L}_{i 2 t}=K L\left(\mathbf{p}_{\mathbf{i}} \| \mathbf{q}_{\mathbf{i}}\right)=\frac{1}{N} \sum_{i=1}^N \sum_{j=1}^N p_{i, j} \log \left(\frac{p_{i, j}}{q_{i, j}+\epsilon}\right),
\end{equation}
where $p_{i,j}$ is the probability denoting the similarity between image-text pairs and $q_{i,j}$ is the true matching probability.
Considering the SDM loss from text to image, the bi-directional SDM loss is formulated as:
\begin{equation} 
\mathcal{L}_{s d m}=\mathcal{L}_{i 2 t}+\mathcal{L}_{t 2 i}.
\end{equation}

Our framework is trained in an end-to-end manner, and the overall optimization objective incorporating auxiliary loss in Equation (\ref{eq:aux_loss}) from vision and language is defined as:
\begin{equation}
\mathcal{L}=\mathcal{L}_{sdm} + \alpha \cdot \left(\mathcal{L}_{a u x}^I+\mathcal{L}_{a u x}^T\right)
\end{equation}

During inference, the trained network incorporating DM-Adapter calculates the similarities between text and image embeddings.
The Top-K candidates are then processed to derive the relevant evaluation metrics for each query.

\section{Experimental Results}
\subsection{Experimental Setup}
\subsubsection{Datasets.} \textbf{CUHK-PEDES} \cite{Li_2017_CVPR} as the most commonly used dataset, contains 40,206 images and 80,412 textual descriptions for 13,003 identities. The training set consists of 11,003 identities with 34, 054 images and 68, 126 texts.
Both the validation set and test set have 1,000 identities.
\textbf{ICFG-PEDES} \cite{ding2021semantically} contains 54,522 images for 4,102 identities. Each image corresponds to one description. The training and test sets contain 3,102 identities and 1,000 identities respectively.
\textbf{RSTPReid} \cite{zhu2021dssl} as a newly released dataset contains 20,505 images of 4,101 identities. Each image has 2 descriptions. The training, validation and test sets contain 3701 identities with 18505 images, 200 identities with 1000 images, and 200 identities with 1000 images respectively.

\subsubsection{Evaluation Measures.} Rank-k metrics (k=1,5,10) are adopted as the primary evaluation metrics, which denote the probability of finding at least one person image matching within the Top-K candidates when given a textual description. 
Additionally, we adopt the mean Average Precision (mAP) as a comprehensive retrieval criterion. 
The higher Rank-k, mAP indicates better performance.

\subsubsection{Implementation Details.}
The framework consists of a pre-trained image encoder, \emph{i.e.}, CLIP-ViT-B/16, a pre-trained text encoder, \emph{i.e.}, CLIP text Transformer, and PETL modules with mixture-of-adapters.
The image is resized to 384 × 128, and the length of textual token sequence is 77.
The model is trained using Adam optimizer for 60 epochs, with a batch size of 128 and an initial learning rate $3 \times {10^{ - 4}}$.
We utilize the reduction parameter 8 representing the bottleneck dimension in adapter as CSKT \cite{liu2024clip}.
Top-K is set to 2, and the number of experts is 6.
The hyperparameter $\alpha$ that indicates the auxiliary loss is set to 0.5.
We perform experiments on a single NVIDIA 4090 24GB GPU.

\subsection{Performance and Memory Efficiency}

We categorize existing methods into those based on CLIP and those based on other architectures.
The primary baseline is recent PETL-based method CSKT \cite{liu2024clip} based on CLIP backbone.

\subsubsection{Results on CUHK-PEDES.}
\begin{table*}[tb]
\small
\centering
\tabcolsep=5pt
\renewcommand\arraystretch{1.1}
\resizebox{0.85\textwidth}{!}{%
\begin{tabular}{c|l|cccccccc}
\hline
                                               & Method  & Type & Ref & Image Enc. & Text Enc. & R@1  & R@5  & R@10   & mAP  \\
\hline
\multirow{9}{*}{\rotatebox{90}{w/o CLIP}}      
                                            & CMPM/C \cite{zhang2018deep}   &  L & ECCV18 & RN50 & LSTM & 49.37     & 71.69      & 79.27      & -        \\
                                               & ViTAA \cite{wang2020vitaa}  &  L & ECCV20 & RN50 & LSTM    & 55.97     & 75.84      & 83.52      & -      \\
                                               & NAFS \cite{gao2021contextual}   & L &arXiv21  & RN50 & BERT & 59.36 & 79.13 & 86.00 & 54.07 \\
                                               & DSSL \cite{zhu2021dssl}   &  L   & MM21 & RN50 & BERT     & 59.98     & 80.41      & 87.56      & -      \\

                                               & SSAN \cite{ding2021semantically} &  L   & arXiv21 & RN50 & LSTM      & 61.37     & 80.15      & 86.73      & -      \\   
                                               & SAF \cite{li2022learning}   &  L &ICASSP22 & ViT-Base & BERT    & 64.13     & 82.62      & 88.40      & 58.61  \\
                                               & TIPCB \cite{chen2022tipcb}  &  L &Neuro22 & RN50 & BERT   & 64.26     & 83.19      & 89.10      & -      \\
                                               & AXM-Net \cite{farooq2022axm}  &  L & MM22 & RN50 & BERT    & 64.44     & 80.52      & 86.77      & 58.73  \\
                                               & LGUR \cite{shao2022learning}  &  L & MM22 & DeiT-Small & BERT & 65.25     & 83.12      & 89.00      & -  \\
                                               & IVT \cite{shu2022see}    & G     & ECCV22 &ViT-Base & BERT   & 65.59     & 83.11      & 89.21      & -  \\

\hline\hline
\multirow{6}{*}{\rotatebox{90}{w/ CLIP}}        & Han et al. \cite{han2021text} & G  &BMVC21 & CLIP-RN101 & CLIP-Transformer   & 64.08     & 81.73      & 88.19      & 60.08  \\

                                               & CFine \cite{yan2023clip}   & L   &TIP23& CLIP-ViT & BERT    & 69.57     & 85.93      & 91.15      & -  \\
                                               & IRRA-CLIP \cite{jiang2023cross} & G &CVPR23 & CLIP-ViT & CLIP-Transformer & 68.19     &86.47      & 91.47      & 61.12           \\
                                                & IRRA$^*$ \cite{jiang2023cross} & G &CVPR23 & CLIP-ViT & CLIP-Transformer & 71.15 & 87.66 & 92.58 & 64.84
                                                \\ 
                                             & IRRA  \cite{jiang2023cross} & G &CVPR23 & CLIP-ViT & CLIP-Transformer & 73.38 & 89.93 & 93.71 & 66.13 \\ 
                                                \cline{2-10} 
                                                
                                               & \cellcolor{gray!40} CSKT \cite{liu2024clip}  & \cellcolor{gray!40} \underline{\textbf{P}+G} &\cellcolor{gray!40} ICASSP24   & \cellcolor{gray!40} CLIP-ViT & \cellcolor{gray!40} CLIP-Transformer & \cellcolor{gray!40} 69.70    & \cellcolor{gray!40} 86.92    & \cellcolor{gray!40} 91.80    & \cellcolor{gray!40} 62.74 \\ 
                                               &   \cellcolor{gray!40} \textbf{DM-Adapter (Ours)}   &  \cellcolor{gray!40} \underline{\textbf{P}+G} & \cellcolor{gray!40} - & \cellcolor{gray!40} CLIP-ViT & \cellcolor{gray!40} CLIP-Transformer & \cellcolor{gray!40} 72.17 & \cellcolor{gray!40}  88.74 & \cellcolor{gray!40} 92.85 & \cellcolor{gray!40} 64.33 \\
\hline
\end{tabular}}

$*$ indicates our replication results after a minor bug correction, which can be regarded as a data augmentation technique in vanilla IRRA.\\
\caption{Comparison Performance with other methods on CUHK-PEDES. The left column denotes whether using  CLIP. ``G'' and ``L'' in ``Type'' denote global and local matching. ``P'' stands for the PETL-related methods (such as CSKT and ours). }
 \label{tab:cuhk}
\end{table*}

As shown in Table \ref{tab:cuhk}, on the most common benchmark CUHK-PEDES, our DM-Adapter outperforms the PETL-based method  CSKT across three Rank-k metrics and mAP by a large margin, with +2.47\%, +1.82\%, +1.05\% and +2.16\%, respectively .
Meanwhile, as shown in Table \ref{tab:effective}, DM-Adapter is comparable with  IRRA, given that IRRA with 195M trainable parameters, integrates a complex implicit reasoning module and sophisticated loss functions. 
In contrast, DM-Adapter with only a few 16M trainable parameters, achieves the trade-off between performance and costs.


\begin{table}[!tb]
    \centering
    \renewcommand\arraystretch{1.1} 
    \resizebox{0.45\textwidth}{!}{%
        \begin{tabular}{lcccccc}
            \hline
            Method  & R@1 $\uparrow$ & Memory Cost (M) $\downarrow$ & Trainable \# Param (M) $\downarrow$ \\
            \hline
            IRRA $^*$      & 71.15 & 7034 (28.64\%)   & 195M \\
            IRRA      & 73.38 & 7034 (28.64\%)   & 195M \\
            CFine      & 69.57 & 13570 (55.24\%)  & 205M \\
            \cline{1-4}
            \cellcolor{gray!40}CSKT       &\cellcolor{gray!40} 69.70 & \cellcolor{gray!40}2338 (9.52\%)   &\cellcolor{gray!40} 12M \\
           \cellcolor{gray!40} \textbf{DM-Adapter (Ours)} &\cellcolor{gray!40} 72.17 & \cellcolor{gray!40} 2952 (12.02\%)  & \cellcolor{gray!40}16M \\ 
            \hline
        \end{tabular}
    }
     \caption{Analysis of Memory Efficiency and Effectiveness on CUHK-PEDES. To ensure a fair comparison of efficiency, batch size for all methods is set to 32.}
    \label{tab:effective}
\end{table}

\subsubsection{Results on  ICFG-PEDES and RSTPReid.}

\begin{table}[tb]
\small
\centering
\tabcolsep=5pt
\renewcommand\arraystretch{1.1}

\resizebox{0.47\textwidth}{!}{%
\begin{tabular}{c|l|cccc}
\hline
                                               & Method    & R@1  & R@5  & R@10   & mAP \\
\hline
\multirow{7}{*}{\rotatebox{90}{w/o CLIP}}                                                                                           & CMPM/C \cite{zhang2018deep}      & 
                                              43.51     & 65.44     & 74.26      & -  \\
                                               & ViTAA \cite{wang2020vitaa}       & 50.98     & 68.79     & 75.78      & -      \\
                                               & SSAN \cite{ding2021semantically} & 54.23     & 72.63     & 79.53      & -      \\
                                               & SAF \cite{li2022learning}        & 54.86     & 72.13     & 79.13      & 32.76    \\
                                               & TIPCB \cite{chen2022tipcb}       & 54.96     & 74.72     & 81.89      & -      \\
                                             & IVT \cite{shu2022see}            & 56.04     & 73.60     & 80.22      & -  \\
                                               & LGUR \cite{shao2022learning}     & 59.02     & 75.32     & 81.56      & -  \\

\hline\hline
\multirow{5}{*}{\rotatebox{90}{w/ CLIP}}
                                               & CFine \cite{yan2023clip}         & 60.83    & 76.55      & 82.42      & -  \\
                                               & IRRA-CLIP \cite{jiang2023cross} & 56.74     & 75.72      & 82.26      & 31.84           \\
                                               & IRRA$^*$ \cite{jiang2023cross}   &  61.36  & 78.66 &84.60 & 37.95           \\ 
                                               & IRRA \cite{jiang2023cross}   &  63.46  & 80.25 &85.82 & 38.06\\ 
                                               \cline{2-6} 
                                               & \cellcolor{gray!40} CSKT \cite{liu2024clip}      & \cellcolor{gray!40} 58.90    &\cellcolor{gray!40} 77.31    & \cellcolor{gray!40} 83.56    & \cellcolor{gray!40} 33.87 \\ 
                                               & \cellcolor{gray!40} \textbf{DM-Adapter (Ours)}     & 
                                               \cellcolor{gray!40} 62.64    &\cellcolor{gray!40} 79.53    & \cellcolor{gray!40} 85.32    &\cellcolor{gray!40} 36.50 \\      

\hline
\end{tabular}}
\caption{Comparison on ICFG-PEDES.}
\label{tab:icfg}
\end{table}

\begin{table}[t]
\small
\centering
\tabcolsep=5pt
\renewcommand\arraystretch{1.1}
\resizebox{0.46\textwidth}{!}{%
\begin{tabular}{c|l|cccc}
\hline
                                               & Method    & R@1  & R@5  & R@10   & mAP \\
\hline
\multirow{4}{*}{\rotatebox{90}{w/o CLIP}}       & DSSL \cite{zhu2021dssl}          & 32.43                                                  & 55.08      & 63.19      & -      \\
                                               & SSAN \cite{ding2021semantically} & 43.50     & 67.80      & 77.15      & -      \\
                                               & SAF \cite{li2022learning}        & 44.05     & 67.30      & 76.25      & 36.81    \\
                                               & IVT \cite{shu2022see}            & 46.70     & 70.00     & 78.80      & -  \\

\hline\hline
\multirow{5}{*}{\rotatebox{90}{w/ CLIP}}       
                                               & CFine \cite{yan2023clip}         & 50.55     & 72.50     & 81.60      & -  \\
                                               & IRRA-CLIP \cite{jiang2023cross} & 
                                               54.05     & 80.70     & \underline{88.00}	& 43.41        \\
                                               & IRRA$^*$ \cite{jiang2023cross} & 57.50 & 80.15 & 87.05 & 44.31       \\ 
                                               & IRRA \cite{jiang2023cross} & 60.20 & 81.30 & 88.20 & 47.17       \\                                      
                                                \cline{2-6} 
                                                
                                               & \cellcolor{gray!40}  CSKT \cite{liu2024clip}    & \cellcolor{gray!40} 57.75    &\cellcolor{gray!40}  81.30   & \cellcolor{gray!40} 88.35    & \cellcolor{gray!40} 46.43 \\ 
                                               & \cellcolor{gray!40}  \textbf{DM-Adapter (Ours)}   & \cellcolor{gray!40} 60.00 & \cellcolor{gray!40} 82.10 & \cellcolor{gray!40} 87.90 &\cellcolor{gray!40}  47.37 \\ 

\hline
\end{tabular}}
\caption{Comparison on RSTPReid.}
\label{tab:rstp}
\end{table}

We then perform experiments on the ICFG-PEDES dataset in Table \ref{tab:icfg} and the newly released RSTPReid dataset in Table \ref{tab:rstp}, which demonstrate the similar comparison results to those in CUHK-PEDES when DM-Adapter is compared with IRRA and CSKT.
DM-Adapter exceeds CSKT by +2.25\% on R@1, +0.8\% on R@5, and +0.94\% on mAP.
Furthermore, we observe that for the most complex CLIP-based method CFine, which incorporates multiple explicit granularity modules, has the worse overall performance on RSTPReid.
DM-Adapter outperforms CFine by an absolute margin of +9.45\% on R@1.
We infer that the full-model model with a larger number of training parameters CFine \cite{yan2023clip} is prone to overfitting on the smallest RSTPReid dataset, compared to the PETL-based methods such as DM-Adapter and CSKT, resulting in poor generalization.

Overall, DM-Adapter reliably delivers the better trade-off between performance and computation costs across all three benchmark datasets, highlighting the generalization and robustness of our proposed approach.

\subsection{Ablation Study}
A comprehensive ablation study for components of DM-Adapter is presented in Table \ref{tab:ablate}, including the most critical accuracy metric R@1 and the average metric on CUHK-PEDES, ICFG-PEDES and RSTPReid datasets.
The results in No.1 serve as the backbone baseline by zero-shot CLIP, where inference is performed directly on the original frozen CLIP model without adding any additional trainable modules.
In No.2, the sub-module adapter in CSKT based on the full-model frozen CLIP for TPR task is reproduced by spanning single adapter on MLP layer of CLIP, and the R@1 metric is enhanced to 71.00\% by following the proper initialization \cite{gao2023unified}.

To demonstrate the effectiveness of our proposed Sparse Mixture-of-Adapters (SMA), we compare it with single adapter (No.2 \emph{vs.} No.3), and show that SMA achieves a significant improvement by +1.00\% on CUHK-PEDES and +0.5\% on RSTPReid.
It indicates that adopting MOE structure enhances capabilities of individual expert, enabling the more fine-grained feature extraction, where each expert processes input tokens from different specialized perspectives.

To further validate the effectiveness of Load-Balancing loss, we compared the performance of SMA trained with (No.3) and without (No.2) the LB loss.
It is evident that LB is valid in alleviating load imbalance, especially on smaller datasets such as RSTPReid.

Moreover, we compare the performance of the model whether using our proposed Domain-Aware Router (DR) or original router (No.3 \emph{vs.} No.4).
The results show that DR can prompt the router to select experts by supplementing domain-aware information and routing, and thus achieve better performance.

In summary, sparse mixture-of-adapters enhances model capacity and helps to unleash the fine-grained feature extraction power of the pretrained CLIP.
It further pushs parameter-efficient transfer learning to the limit for the specific TPR task with domain information prompts.
Compared to the original adapter MLP-Adapter (No.2), DM-Adapter (No.4) achieves significant improvements across all three datasets, gaining a substantial increment by +1.05\% on the overall average R@1 metric.

\begin{table*}[!tb]
\centering
\tabcolsep=3.5pt
\renewcommand\arraystretch{1.1} 
  \resizebox{0.85\textwidth}{!}{%
  \begin{tabular}{c|l|cccc|c|c|c|c}
  \toprule
  \multirow{2}{*}{No.} &\multirow{2}{*}{Methods} &\multicolumn{4}{c|}{Components} &\multirow{2}{*}{CUHK-PEDES} & \multirow{2}{*}{ICFG-PEDES} &  \multirow{2}{*}{RSTPReid} &  \multirow{2}{*}{Avg.}
  \\ 
  \cline{3-6}
       &     &MLP-Adapter & SMA     &LB   & DR    &   &  &   &  \\ 
       
  \hline
 0    &Zero-shot CLIP                               &    &      &          &          &12.65	 &6.66	   &13.55 	&10.96    \\
  1    & + MLP-Adapter  & \checkmark   &  &  & & 71.00   & 62.13 &  58.55& 63.89 \\
  2    & + Sparse Mixture-of-Adapters (w/o LB) & & \checkmark & & & 72.00  & 62.13 & 59.05 & 64.38    \\
  3    & + Sparse Mixture-of-Adapters ((w/ LB))  & &\checkmark  &\checkmark &    &\underline{72.14} & \underline{62.40}  & \underline{59.40}  &  \underline{64.65} \\    
  4    & + Domain-Aware Router (\textbf{DM-Adapter})  & &\checkmark   &\checkmark & \checkmark &   \textbf{72.17}   & \textbf{62.64} &  \textbf{60.00} & \textbf{64.94} \\ 
  \bottomrule
  \end{tabular}%
  }
  \caption{Ablation study on R@1 about each component of DM-Adapter. The metric Avg. denotes the average R@1 across three datasets. We reproduce MLP-Adapter \cite{liu2024clip} by following the proper initialization \cite{gao2023unified}.}
  \label{tab:ablate}
  \end{table*}



\subsection{Hyper-parameter Analysis}

\subsubsection{The Number of Experts.}
As shown in Figure \ref{fig: moe_hyper} (\emph{upper}), to investigate the impact of the number of experts $n$, we sample $n$ as 2, 4, 6, 8 and 10 to evaluate the R@1 and trainable parameters under different numbers of experts. 
A constant Top-K value is fixed to 2 in the overall experiment.
When $n$ is less than 6, average R@1  gradually increases with an increasing number of experts. 
However, when $n$ exceeds 6, larger $n$ leads to a decrease in performance.
we observe that although increasing $n$ can proportionally enhance the model's information capacity, a larger $n$ does not necessarily lead to better performance.
This suggests that the model's capacity cannot grow indefinitely and should be aligned with the scale of the training data.
We ultimately determine $n=6$ as a practical choice.

\subsubsection{The Number of Top-K.}
\begin{figure}[]
	\centering
	\subfloat{
		\includegraphics[width=0.82\linewidth]{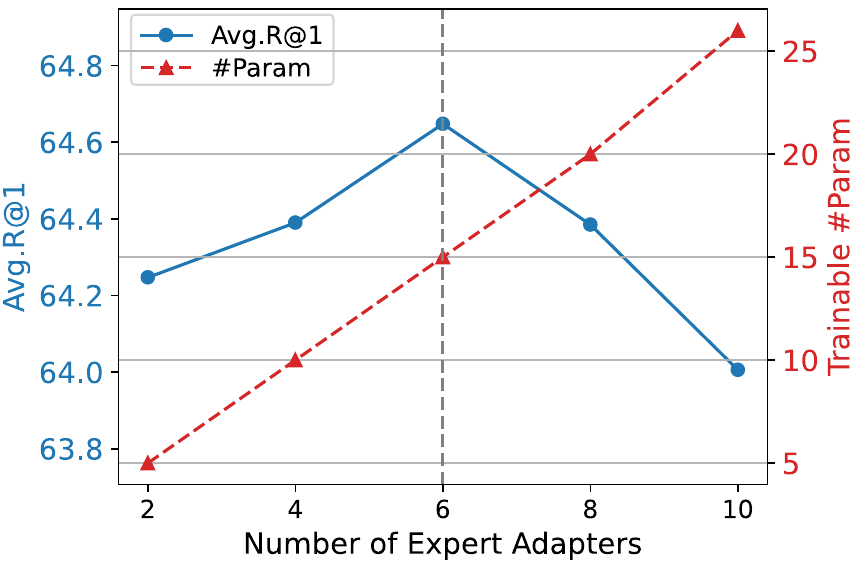}} \\ \quad 
	\subfloat{
		\includegraphics[width=0.82\linewidth]{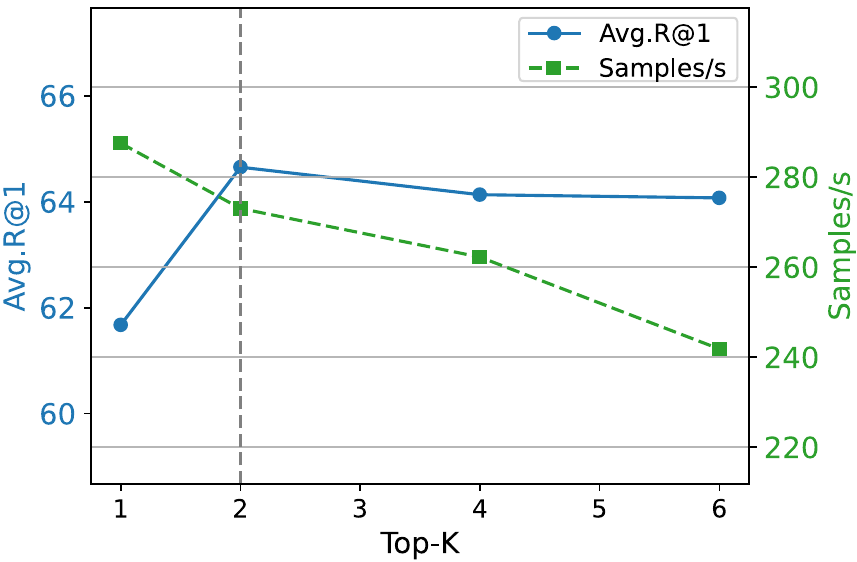}}
	\caption{\normalsize{The results of experiments for hyper-parameters.} 
	}
	\label{fig: moe_hyper}
\end{figure}

In Figure \ref{fig: moe_hyper} (\emph{lower}), we set the number of experts $n=6$ and explore the R@1 accuracy and computational complexity with the change of Top-K.
It clearly demonstrates that the sample processing speed deteriorates as Top-K increases.
When Top-K is 1, performance shows worse as it would damage the powerful ability of mixture-of-experts and degrades to a single adapter.
As Top-K increases further, the model's performance essentially reaches a plateau without additional improvement.
Given the need to strike a balance between efficiency and performance, a practical choice for Top-K would be 2.

\subsection{Visualization of Expert Weight}
\begin{figure}[tb]
    \centering
    \includegraphics[width=0.93\linewidth]{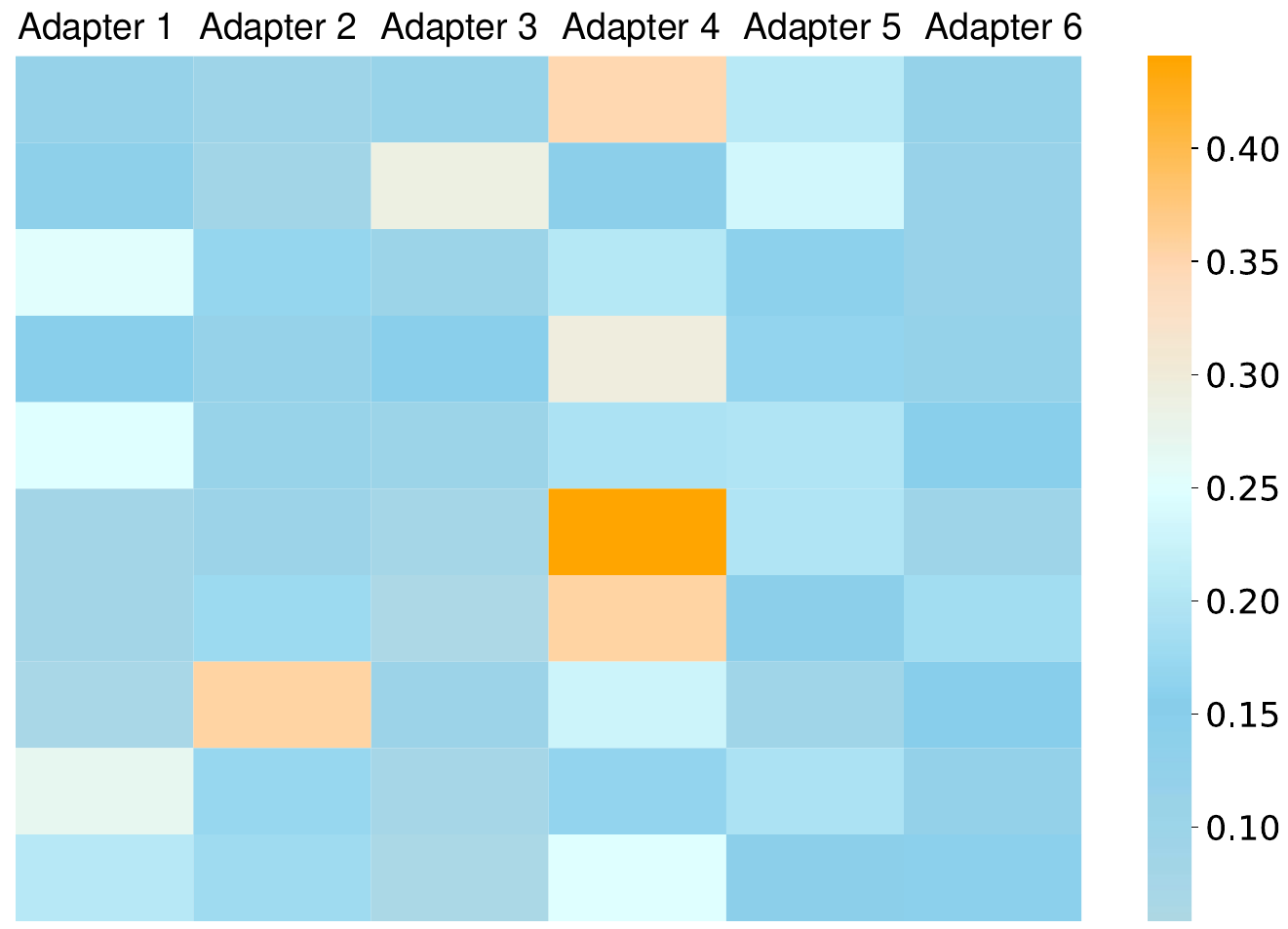}
    \caption{\normalsize{Visualization of Expert Weight. Each row represents the weights assigned to different experts for tokens of a description. Each column indicates a expert adapter. }}
\label{fig: visual}
\end{figure}

In Figure \ref{fig: visual}, we present a weight visualization of the 6 expert adapters for a person description.
We analyze the weights of DM-Adapter in the 12\emph{th} layer of CLIP, as it represents the most high-level features.
Each row represents the distribution of weights assigned to  input tokens across mixture-of-experts, and the sum of the weights equals 1 under normalization.
This demonstrates that DM-Adapter can implicitly handle feature granularity more precisely, as different experts specialize in distinct aspects of person knowledge.

\section{Conclusion}
In this paper, we present a novel CLIP-based parameter-efficient transfer learning method DM-Adapter to achieve implicit fine-grained knowledge transferring, which freezes the entire CLIP backbone and only trains a few parameters with domain-aware mixture-of-adapters.
Extensive experiments demonstrate that our approach achieves the best trade-off between robust performance and parameter efficiency, and outperforms existing the PETL-based methods.

\section{Acknowledgements} 
\label{sec:acknowledge}
This work was partly supported by the Special Foundations for the Development of Strategic Emerging Industries of Shenzhen(No.KJZD20231023094700001) and National Natural Science Foundation of China (62402252).

\bibliography{aaai25}

\end{document}